\documentclass[manuscript,screen,nonacm,natbib=false]{acmart}

\usepackage{multirow}
\usepackage{booktabs}
\usepackage{tabularx}
\usepackage{caption}
\usepackage{lineno}

\usepackage{graphicx}%
\usepackage{multirow}%
\usepackage{amsmath,amsfonts}%
\usepackage{amsthm}%
\usepackage{mathrsfs}%
\usepackage[title]{appendix}%
\usepackage{xcolor}%
\usepackage{textcomp}%
\usepackage{manyfoot}%
\usepackage{booktabs}%
\usepackage{algorithm}%
\usepackage{algorithmicx}%
\usepackage{algpseudocode}%
\usepackage{listings}%
\usepackage{printlen}
\usepackage{hyperref}
\usepackage{soul}
\usepackage{siunitx}
\newcommand{\makesingular}[2]{\expandafter\newcommand\csname #1\endcsname{#2}}
\newcommand{\makeplural}[2]{\expandafter\newcommand\csname #1s\endcsname{#2s}}

\newcommand{\newWord}[2]{%
    \makesingular{#1}{#2}
    \makeplural{#1}{#2}
}

\newWord{methodName}{Neural Jacobian Field}

\newcommand{\robotconfig}[1]{\mathbf{q}}
\newcommand{\robotcommand}[1]{\mathbf{u}}

\newcommand{\imageInput}[1]{\mathbf{I}}

\newcommand{\coord}{\ensuremath{\textbf{x}}}
\newcommand{\R}{\ensuremath{\mathbb{R}}}
\newcommand{\volrendidx}{\ensuremath{s}}
\newcommand{\joints}{\ensuremath{\textbf{u}}}
\newcommand{\desired}{\ensuremath{\textbf{v}}}
\newcommand{\oflow}{\ensuremath{\textbf{v}}}
\newcommand{\depth}{\ensuremath{\textbf{d}}}
\newcommand{\ray}{\ensuremath{\textbf{r}}}
\newcommand{\pose}{\ensuremath{\textbf{P}}}
\newcommand{\intrinsics}{\ensuremath{\textbf{K}}}
\newcommand{\img}{\ensuremath{\textbf{I}}}
\newcommand{\h}{\text{H}}
\newcommand{\w}{\text{W}}
\newcommand{\featvol}{\textbf{W}}
\newcommand{\jacobian}{\textbf{J}}
\newcommand{\density}{\ensuremath{\sigma}}
\newcommand{\radiance}{\textbf{c}}

\newcommand{\du}{\ensuremath{\delta \joints}}
\newcommand{\dx}{\ensuremath{\delta \coord}}
\newcommand{\jf}{\ensuremath{\mathbf{J}(\coord, \imageInput{})}}

\AtBeginDocument{%
  }

\setcopyright{acmlicensed}
\copyrightyear{2018}
\acmYear{2018}
\acmDOI{XXXXXXX.XXXXXXX}

\acmConference[Conference acronym 'XX]{Make sure to enter the correct
  conference title from your rights confirmation emai}{June 03--05,
  2018}{Woodstock, NY}
\acmISBN{978-1-4503-XXXX-X/18/06}

\RequirePackage[
datamodel=acmdatamodel,
style=acmnumeric, 
]{biblatex}
\begin{document}

\title{Controlling Diverse Robots by Inferring Jacobian Fields with Deep Networks}

\author{Sizhe Lester Li}
\email{sizheli@mit.edu}
\authornotemark[1]

\author{Annan Zhang}
\email{zhang@csail.mit.edu}

\author{Boyuan Chen}
\email{boyuanc@mit.edu}

\author{Hanna Matusik}
\email{hania@csail.mit.edu}

\author{Chao Liu}
\email{chaoliu@csail.mit.edu}

\author{Daniela Rus}
\email{rus@csail.mit.edu}

\author{Vincent Sitzmann}
\email{sitzmann@mit.edu}
\authornote{Corresponding authors: \href{mailto:sizheli@mit.edu}{sizheli@mit.edu}, \href{mailto:sitzmann@mit.edu}{sitzmann@mit.edu}. All authors are with the Computer Science and Artificial Intelligence Laboratory (CSAIL), Massachusetts Institute of Technology, 32 Vassar St, Cambridge, MA 02139, USA.}


\renewcommand{\shortauthors}{S. L. Li, A. Zhang, B. Chen, H. Matusik, C. Liu, D. Rus, V. Sitzmann}

\begin{abstract}
{Mirroring the complex structures and diverse functions of natural organisms is a long-standing challenge in robotics~\cite{yang2018grand,kim2013soft,trivedi2008soft,pfeifer2007self}.
Modern fabrication techniques have dramatically expanded feasible hardware~\cite{majidi2019soft,buchner2023vision,mcevoy2015materials,evenchik2023electrically}, yet deploying these systems requires control software to translate desired motions into actuator commands.
While conventional robots can easily be modeled as rigid links connected via joints, it remains an open challenge to model and control bio-inspired robots that are often multi-material or soft, lack sensing capabilities, and may change their material properties with use~\cite{rus2015design,armanini2023soft,wang2022control,della2023model}.
Here, we introduce a method that uses deep neural networks to map a video stream of a robot to its Visuomotor Jacobian Field: the sensitivity of all 3D points to the robot's actuators. Our method enables controlling robots from only a single camera, makes no assumptions about the robots' materials, actuation, or sensing, and is trained without expert intervention by observing the execution of random commands.
We demonstrate our method on a diverse set of robot manipulators, varying in actuation, materials, fabrication, and cost.
Our approach achieves accurate closed-loop control and recovers the causal dynamic structure of each robot.
By enabling robot control with a generic camera as the only sensor, we anticipate our work will broaden the design space of robotic systems and serve as a starting point for lowering the barrier to robotic automation.}
Additional materials can be found on the project website \footnote{\url{https://sizhe-li.github.io/publication/neural_jacobian_field}}.

\end{abstract}





\maketitle

\section{Introduction}
\label{sec:introduction}
Modern manufacturing techniques promise a new generation of robotic systems inspired by the diverse mechanisms seen in nature.
While conventional systems are precision engineered from rigid parts connected at discrete joints, bio-inspired robots generally combine soft, compliant materials and rigid parts, and often forego conventional motor-driven actuation for pneumatic and muscle-like actuators~\cite{rus2015design}.
Recent work has demonstrated that such hybrid soft-rigid systems can already outperform conventional counterparts in certain environments where adaptation to changing circumstances~\cite{laschi2016soft,pratt1995series} or safety in co-working with humans is key~\cite{althoff2019effortless}.
Further, these systems are amenable to mass production, some requiring no human assembly~\cite{buchner2023vision}, and may thus dramatically lower the cost and barriers to robotic automation~\cite{li2017fluid}.
However, deployment of bio-inspired hardware is hindered by our capability of modeling these systems: Any robotic system needs to be paired with a model that can accurately predict the motion of key components, such as the end-effector, under all possible commands at all times.

Conventional robots were designed to make their modeling and control easy. They are usually constructed from precision-machined parts fabricated out of high-stiffness materials with Young's moduli in the $\SI{e9}-\SI{e12}{\pascal}$ range~\cite{rus2015design}. Connected by low-tolerance joints, these rigid robots are adequately modeled as a kinematic chain consisting of idealized rigid links.
Accurate sensors in every joint then allow a faithful 3D reconstruction of the robot during deployment.
With this in place, an expert can reliably model the motion of the robot under all possible motor commands, and design control algorithms to execute desired motions.

In contrast, the bodies of soft and bio-inspired robots are difficult to model. They are typically made out of materials that match the stiffness of soft biological materials such as tissue, muscles, or tendons~\cite{rus2015design,majidi2019soft}. 
These materials undergo large deformations during actuation and exhibit time-dependent effects such as viscoelasticity and gradual weakening through repeated loading and unloading.
Partial differential equations that govern the behavior of soft materials derived from continuum mechanics and large deformation theory are costly to solve, especially for control and real-time applications. Model order reduction methods, geometrical approximation methods, and rigid discretization methods rely heavily on simplifying assumptions about the specific system and do not universally generalize~\cite{della2023model,armanini2023soft,wang2022control}.
Prior work has leveraged machine learning~\cite{chin2020machine,kim2021review,george2021machine, truby2020distributed, thuruthel2019soft} and marker-based visual servoing~\cite{lagneau2020active, yip2014model, albeladi2022hybrid, bern2020soft, hosoda1994versatile} to overcome these challenges, but requires extensive expert-guided customization to be applied to a particular robot architecture. 
Ref.~\cite{toshimitsu2022dije} estimates the 2D image Jacobian online using a history of optical flow and a Kalman filter. 
The method is markerless and can visually control a low-DOF robot arm. However, it does not learn from visual observations or use 3D structure, making it unsuitable when the robot leaves the frame, rotates, deforms significantly, or becomes occluded, as noted by the authors.

%

%
Further, high-precision motion capture systems~\cite{optitrack, vicon, qualisys} are costly, bulky, and require a controlled setting for deployment.
Recent work explores neural scene representations of robot morphology~\cite{chen2022fully}, but assumes precise embedded sensors unavailable in soft and bio-inspired robots, and relies on 3D motion-capture for fine-grained control.
What is required is a general-purpose control method that is agnostic to the fabrication, actuation, embedded sensors, material, and morphology of the robotic system.
\begin{figure}[t!]
\centering
\includegraphics[width=\textwidth]{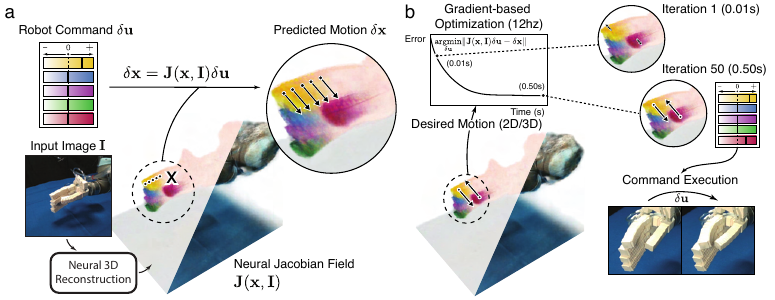}
\caption{
\textbf{Controlling robots from vision via the Neural Jacobian Field.} 
\textbf{a,} Reconstruction of the Neural Jacobian Field and motion prediction. From a single image, a machine learning model infers a 3D representation of the robot in the scene, the Neural Jacobian Field. It encodes the robot's geometry and kinematics, allowing us to predict the 3D motions of robot surface points under all possible commands. Coloration indicates the sensitivity of that point to individual command channels.
\textbf{b,} Closed-loop control from vision. Given desired motion trajectories in pixel space or in 3D, we use the Neural Jacobian Field to optimize for the robot command that would generate the prescribed motion at an interactive speed of approximately 12 Hz. Executing the robot command in the real world confirms that the desired motions are achieved.
}
\label{fig:teaser}
\end{figure}

The work in this article is inspired by human perception. 
Controlling robots with just a video game controller, humans can learn to pick and place objects within minutes~\cite{losey2022learning}.
The only sensors we require are our eyes: From vision alone, we learn to reconstruct the robot’s 3D configuration and to predict its motion as a function of the control inputs we generate.
%

In this article, we introduce \methodNames{}, a machine-learning approach that can control robots from a single video camera stream.
We train our framework using 2-3 hours of multi-view video of the robot executing randomly generated commands captured by twelve consumer-grade RGB-D video cameras.
No human annotation or expert customization is necessary to learn to control a new robot.
After training, our method can control the robot to execute desired motions using only a single video camera.
Relying on vision as the only sensor, \methodNames{} do not make assumptions about the kinematics, dynamics, material, actuation, or sensing capabilities of the robot. 
Our method is uniquely enabled by recent advancements in computer vision, neural scene representation, motion tracking, and differentiable rendering~\cite{zhou2020tracking, tewari2022advances, xie2022neural}.

We evaluate \methodName{} on a wide range of robotic 
manipulation systems, specifically a 3D printed hybrid soft-rigid pneumatic hand~\cite{matusik2023directly}, a compliant wrist-like robotic platform made out of handed shearing auxetics (HSA)~\cite{lipton2018handedness}, a commercially available Allegro hand with 16 degrees of freedom~\cite{allegrohand}, and a low-cost educational robot arm~\cite{lapeyre2015poppy}. Across all these systems, we show that our method reliably learns to reconstruct their 3D configuration and predict their motion at all times.

Our method unshackles the hardware design of robots from our ability to manually model them, which in the past has dictated precision manufacturing, costly materials, extensive sensing capabilities, and reliance on conventional, rigid building blocks.
Our method thus has the potential to dramatically broaden the design space of robots that can be deployed in practice as well as lowering the cost and barriers to adopting robotic automation by enabling precision control of low-cost robots.

\section{The Neural Jacobian Field}
\label{sec:method}

Our framework comprises two key components: 
(1) a deep-learning-based state estimation model that infers a 3D representation of the robot that encodes both its 3D geometry as well as its differential kinematics---how any point in 3D will move under any possible robot command---from only a single video stream.
(2) A closed-loop inverse dynamics controller that translates desired motions specified in the 2D input to robot commands at interactive speeds.
A schematic overview of the deployed system is shown in Figure~\ref{fig:teaser}. 

%
The state estimation model is a deep learning architecture that maps a single image~$\img$ of the robot to a 3D neural scene representation. This 3D representation maps any 3D coordinate to features that describe the robot's geometric and kinematic properties at that 3D coordinate~\cite{mildenhall2020nerf,sitzmann2019siren, xie2022neural}.
Specifically, we reconstruct both a neural radiance field~\cite{mildenhall2020nerf} that encodes the robot's 3D shape and appearance at every 3D coordinate, as well as a novel Neural Jacobian Field that maps each point in 3D to a linear operator that expresses that point's 3D motion as a function of robot actuator commands.

The neural radiance field maps a 3D coordinate to its density and radiance. This serves as a representation of the geometry of the robot, as the density value at every point in the 3D scene encodes the regions of 3D space occupied by the robot.

The Neural Jacobian Field encodes how any 3D coordinate will move as a function of any possible actuator command. It serves as a representation of the differential kinematics of the robot.
It generalizes the conventional \emph{system Jacobian} in robotics.
Traditionally, experts model a robot by designing a dynamical system that has state $\mathbf{q} \in \mathbb{R}^{m}$, input command $\mathbf{u} \in \mathbb{R}^{n}$, and dynamics $\mathbf{q}^{+} = \mathbf{f}(\mathbf{q}, \mathbf{u})$ where $\mathbf{q}^+$ denotes the state of the next timestep. 
The system Jacobian $\jacobian(\mathbf{q}, \mathbf{u}) = \frac{\partial \mathbf{f}(\mathbf{q}, \mathbf{u})}{\partial \mathbf{u}}$ is the matrix that relates the change of command $\mathbf{u}$ to the change of state $\mathbf{q}$, which arises from the linearization of $\mathbf{f}$ around the nominal point $(\mathbf{\bar{q}}, \mathbf{\bar{u}})$, as $\delta \mathbf{q} = \jacobian\vert_{\mathbf{\bar{q}}, \mathbf{\bar{u}}} \delta \mathbf{u}$.
This approach relies on experts to design the system's state encoding $\mathbf{q}$ and dynamics $\mathbf{f}$ on a case-by-case basis. 
While this is feasible for conventional robots, it is challenging for hybrid soft-rigid, insufficiently sensorized, and under-actuated systems, or systems with significant backlash due to imprecise manufacturing.

Our Neural Jacobian Field instead directly maps any 3D point \coord{} to its corresponding system Jacobian. Instead of conditioning on an expert-designed state representation $\mathbf{q}$, the Jacobian Field is reconstructed directly from the input image $\img$ via deep learning~\cite{pixelnerf}.
Specifically, $\jacobian(\coord, \imageInput{}) = \frac{\partial \mathbf{x}}{\partial{\mathbf{u}}}$ describes how a change of actuator state $\du$ relates to the change of 3D motion at coordinate $\mathbf{x}$, via $\dx = \jf \du$. This allows us to densely predict the 3D motion of any point in space using $\dx = \jf \du$. Figure~\ref{fig:teaser}a illustrates the 3D Jacobian field reconstructed by our method to predict the motion of all 3D points belonging to a soft pneumatic hand.

To reconstruct Neural Jacobian and Radiance Fields, we rely on a neural single-view-to-3D module~\cite{pixelnerf}.
%
By reconstructing both robot geometry and kinematics directly from a camera observation, our state estimation model is agnostic to the sensors embedded in the robot. Instead of expert-modeling the relationship between motor and sensor readings and the 3D geometry and kinematics of the robot, our system directly learns to regress this relationship from data.

Our state estimation model is trained self-supervised using video streams from 12 RGB-D cameras that observe the robot executing random commands from different perspectives. We provide a detailed illustration of the training process in Extended Data Figure~\ref{fig:fig5_method_overview}. 
For each camera stream, we extract 2D motion using optical flow and point-tracking methods.
At every training step, we select one of the 12 cameras as input for our reconstruction method. From this single input image, we reconstruct the Neural Jacobian and Radiance Fields that encode the robot's 3D geometry and appearance. Given a robot command, we use the Jacobian Field to predict the resulting 3D motion field. We use volume rendering~\cite{mildenhall2020nerf} to render the 3D motion field to 2D optical flow of one of the other 12 cameras and compare with the observed optical flow. This procedure trains the Jacobian Field to predict robot motion accurately. We further volume render the radiance field from one of the 12 cameras and compare the RGB and depth outputs with the captured RGB-D images, which trains our model to reconstruct accurate 3D geometry. 

After training, our framework enables closed-loop control to execute desired motions on the robot system with a single camera.
Given a set of desired 2D motion vectors $\desired$ specified on robot pixel locations, we use a gradient-based optimizer (Fig.~\ref{fig:teaser}b) to solve for the robot command $\delta\joints{}$ at interactive speeds.
%

The Jacobian Field can be directly integrated into a differential inverse kinematics controller. 
In our experiments, we create reference 2D motion trajectories by recording videos where we teleoperate the robot to execute a desired physical task and extracting 2D point tracks~\cite{doersch2023tapir} from each video that captures the robot's motion in image space.
The controller generates actuator commands that translate to 2D point trajectories that follow the reference trajectory.
The same camera is used to provide a video stream for control. Given the current video frame, we use the point tracker's features to find the most similar point in the reference video for each point in the current video.
The controller then first controls the robot to replicate the configuration observed in the initial frame of the reference trajectory. 
The desired 2D motion is computed as the difference between the target and current 2D locations. The controller then uses the Jacobian Field to search for the robot command $\delta\joints{}$ that could generate the desired motions.
The controller executes the command on the robot, repeats the search, and advances to the next step in the reference trajectory if the L2 norm of the desired motions is under a specified threshold. The control process terminates if the controller has reached the final reference step under the specified threshold.
%
%
%
%
We refer the readers to the Extended Data Section~\ref{app:methods} for details on our method.

\section{Results}
\label{sec:results}
\begin{figure}[thbp!]
\centering
\includegraphics[width=.92\textwidth]{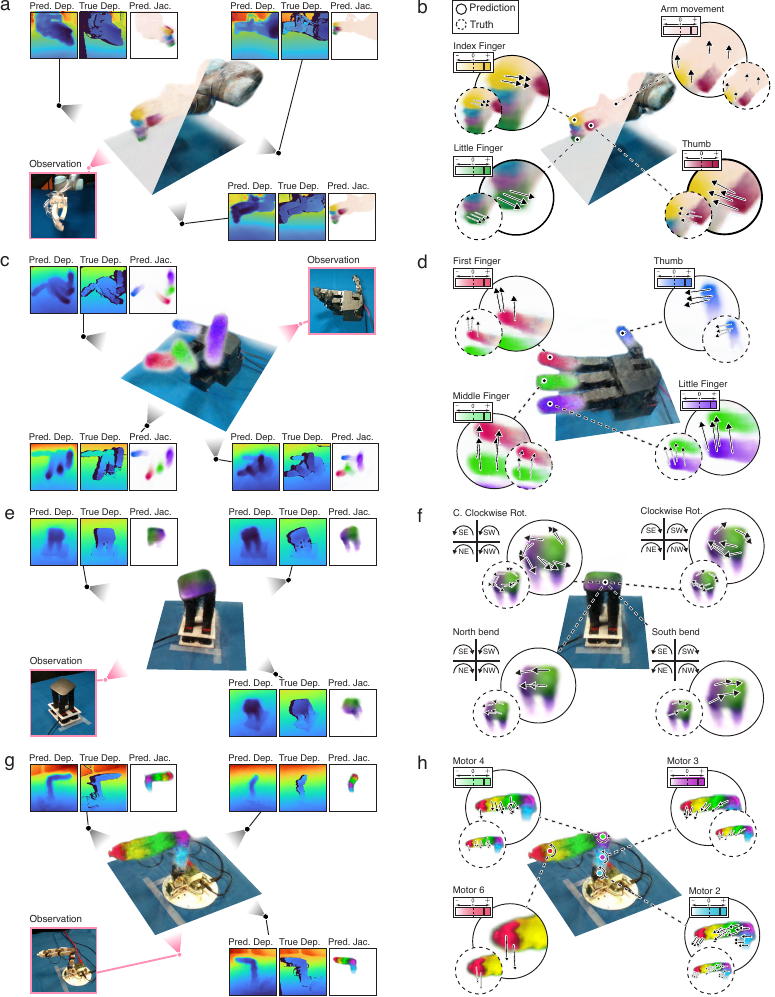}
\vspace{-10pt}
\caption{
\textbf{Reconstruction of robot geometry and kinematics from a single image.} 
\textbf{(a, c, e, g)}, Visualization of the reconstructed Jacobian and Radiance Fields (center) and comparison of reconstructed and measured geometry (sides) from a single input image.  Colorization indicates the motion sensitivity of the 3D point to different actuator command channels, meaning that our system successfully learns correspondence between robot 3D parts and command channels without human annotations. We show depth predictions next to measurements of RGB-D cameras, demonstrating the accuracy of the 3D reconstruction across all systems. 
\textbf{(b, d, f, h)}, 3D motions predicted via the Jacobian Field. We display the motions predicted via the Neural Jacobian Field (solid circle) for various motor commands next to reference motions reconstructed from video streams via point tracking (dotted circle). Reconstructed motions are qualitatively accurate across all robotic systems. Although we manually color-code command channels, our framework associates command channels with 3D motions without supervision.
%
}
\label{fig:fig2_forward_result}
\end{figure}
\begin{figure}[thbp!]
\centering
\includegraphics[width=\textwidth]{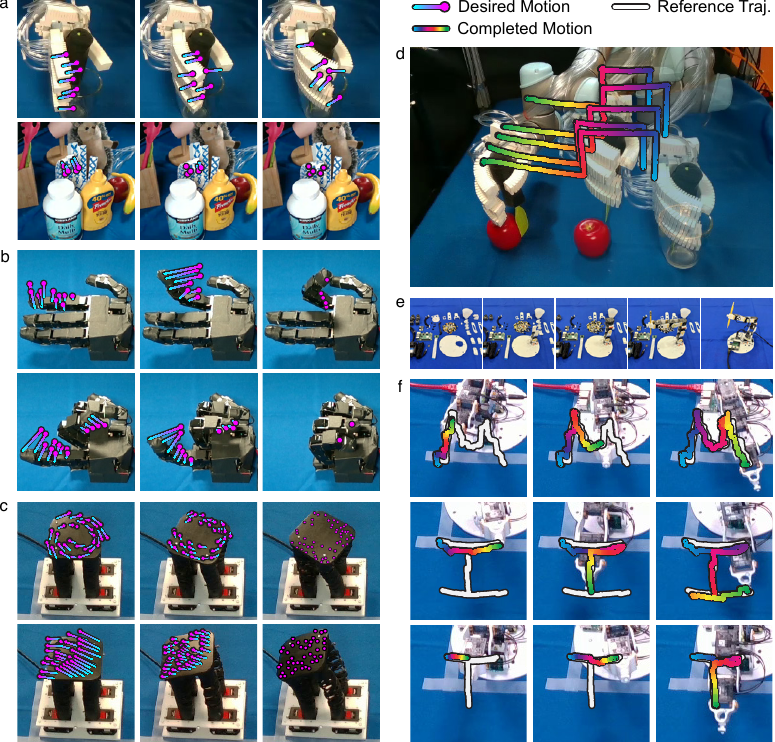}
\caption{
%
\textbf{Closed-loop control of diverse robots from vision.} \textbf{a}, We control a 3D-printed soft-rigid pneumatic hand to complete a grasp (top) and execute finger motion under the presence of occluders (bottom). 
\textbf{b}, Our method precisely controls every finger of the Allegro hand to close and form a fist.
\textbf{c}, We use our system to control complex rotational and bending motions on the wrist-like, soft handed shearing auxetics platform.
\textbf{d}, We control a system that mounts a soft-rigid pneumatic hand on a conventional UR5 robot arm to accomplish a tool grasp and a pushing action. 
\textbf{e}, We show the process of assembling a low-cost, 3D-printed robot arm that is difficult to model and not equipped with any sensors
~\cite{salvato2021characterization, chevalier2019robo, golemo2018sim}.
\textbf{f}, Our method is robust against backlash and jerky motions of the low-cost motors, successfully enabling the robot arm to draw letters ``M'', ``I'', and ``T'' in the air. 
}
\label{fig:fig3_control_result}
\end{figure}
\begin{figure}[thbp!]
\centering
\includegraphics[width=\textwidth]{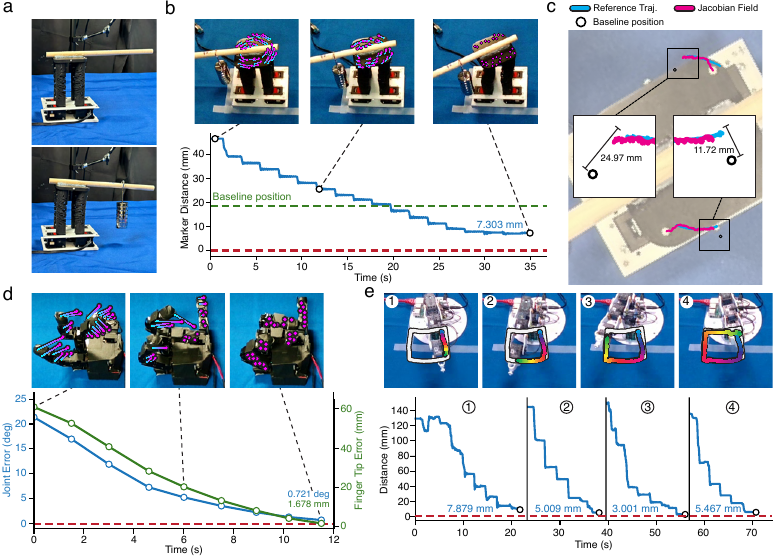}
\caption{
\textbf{Quantitative analysis and resiliency test.}
\textbf{a}, 
We modify the dynamics of the HSA platform.
We attach a rod to the platform and append \SI{350}{\gram} calibration weights at a controlled location, which leads the platform to tilt in its resting position.
%
%
\textbf{b}, Our framework enables the HSA system with changed dynamics to complete the rotation motion (top). We report the distance-from-goal over time (bottom). 
\textbf{c}, Using a bird's-eye view, we overlay the completed 3D trajectory on top of the starting configuration of the HSA platform. We compare the execution trajectory of our approach with the reference trajectory. This visualization confirms that our method is able to counteract the physical effects of the weight and stabilize the motion trajectory toward the target path.
\textbf{d}, We show distance-from-goal on the Allegro Hand, which decreases over time as we execute the motion plan. We measure distance-from-goal using both joint errors in degree and finger-tip positions in millimeters.
\textbf{e}, 
Using a square drawing task, on the top, we visualize the reference trajectory in white and the completed trajectory in rainbow. On the bottom, we plot distance-from-goal over time on the Poppy robot arm in four trajectory segments. 
}
\label{fig:fig4_numerical_result}
\end{figure}

We show the new capabilities enabled by our framework by controlling robotic systems that cover diverse material types, varying kinematic complexity, and different price points.
In summary, we control a $\$300$, 3D-printed hybrid soft-rigid pneumatic hand mounted on a conventional robot arm, a soft parallel manipulator made from handed shearing auxetics, a 16-degrees-of-freedom rigid Allegro hand, and a manually assembled DIY robot arm with 3D-printed parts, low-cost motors, and significant backlash~\cite{golemo2018sim} (Supplementary Video 1).

As shown in Figure~\ref{fig:fig2_forward_result}, for each of these challenging robotic systems, \methodName{} succeeds at reconstructing an accurate 3D representation of the respective robot from just a single image. 
We assign a unique color to the influence of each channel of the $N$-dimensional motor command and visualize the Jacobian Field. We find that our Jacobian Field learns the causal kinematic structure of each robot, identifying which command channel is responsible for actuating which part of the robot in 3D space. This capability arises fully self-supervised, without any annotation or supervision that would match motors with robot parts. 
%
%
We further demonstrate qualitatively (Fig.~\ref{fig:fig2_forward_result}) and quantitatively (Extended Data Table~\ref{tab:table2_perception_result}) that, given a variety of motor commands, the 3D Jacobian field inferred from a single image successfully predicts the motions of 3D points on the robot. 
We qualitatively find that 3D motions predicted by our framework given robot commands highly agree with the ground truth reference motions. (Fig.~\ref{fig:fig2_forward_result}).
Quantitatively, across robotic systems, our method reconstructs high-quality geometry and kinematics from just a single RGB input view. 
The mean depth prediction error is the largest for the pneumatic hand (6.519 mm), and smallest for the HSA platform (1.109 mm).
The pneumatic hand is actuated through translucent tubing that could pose challenges to geometry prediction.
For flow prediction, our framework achieves mean prediction errors of 1.305 pixels and 1.150 pixels on the Allegro hand and pneumatic hand, respectively. The Poppy robot arm has the largest mean flow prediction error of 6.797 pixels, as the hardware constantly experiences backlash due to low-quality motors. Our method is able to model the rotational and bending mechanisms of the HSA platform and achieves a mean flow prediction error of 3.597 pixels.

%
We next evaluate the performance of our method for closed-loop control. For the Allegro hand, we prescribe the controller with a 2D trajectory that tracks a desired pose (Fig.~\ref{fig:fig3_control_result}b). Upon completion of the trajectory, we quantify error using the built-in, high-precision per-joint sensors and the hand's precise 3D forward kinematics model. Purely from vision, our system controls the Allegro hand to close and open every finger fully, achieving errors of less than \ang{3} per joint and less than \SI{4}{\milli\meter} for each fingertip (Fig.~\ref{fig:fig4_numerical_result}d, Extended Data Table~\ref{tab:table1_control_result}, Extended Data Figure~\ref{fig:fig7_more_visual_result}).

We demonstrate on the HSA platform that our system can successfully control robots under dramatically changed dynamics without retraining. We intentionally disturb the HSA platform by attaching calibration weights with a total mass of \SI{350}{\gram} to a wooden rod, which we glue to the top of the HSA platform. The weights exert a vertical force and a torque on the platform top, which makes it tilt visibly in its resting position (Fig.~\ref{fig:fig4_numerical_result}a). Furthermore, the rod and the weights constitute a visual disturbance.
We use the OptiTrack motion capture system~\cite{optitrack} ($<$\SI{0.2}{\milli\meter} of measurement error) and attach markers on the surface of the HSA platform to quantify the position errors in goal pose tracking.
%
%
We find that our vision-based framework is capable of controlling the robot to complete complex rotational motions and reach the target configuration, achieving an error of \SI{7.303}{\milli\meter}, effectively overcoming external perturbation on the system's dynamics (Fig.~\ref{fig:fig4_numerical_result}b,c, Extended Data Table~\ref{tab:table1_control_result}).

For the 3D-printed poppy robot arm, we design target trajectories demanding the robot to draw a square and the letters ``MIT'' in the air. These motion sequences are out-of-distribution and do not exist in our training data. We attach OptiTrack markers on the end-effector of the robot arm to measure 3D position errors. Our framework achieves an average error of less than \SI{6}{\milli\meter} in the goal pose tracking task (Fig.~\ref{fig:fig4_numerical_result}e, Extended Data Table~\ref{tab:table1_control_result}).

Overall, our framework enables precise control of diverse robotic systems, including both conventional, rigid systems as well as 3D-printed, hybrid-material systems, without any expert modeling, intervention, or other per-robot specialization of the algorithm. 
Figure~\ref{fig:fig3_control_result} demonstrates how our system controls the diverse robotic platforms towards executing a variety of skills.
The system achieves smooth trajectories and succeeds at controlling the pneumatic hand mounted on the UR5 robot to pick up a tool from a glass and use it to push an apple.
On the Allegro Hand, our system forms a fist. On the HSA platform, it executes a variety of extension and rotation commands.
Finally, our method is able to control the low-cost poppy robot arm to trace the letters ``MIT''. 
%
To sum up, across a diverse set of robots, our system can control these systems to perform a variety of long-term skills without any expert modeling or customization.
\section{Discussion}
\label{sec:discussion}

We have presented a vision-based deep learning approach that learns to control robots from vision alone, without any assumptions on the robot’s materials, actuation, or embedded sensors.
Across challenging robotic platforms, ranging from conventional rigid systems to hybrid soft-rigid, 3D-printed, compliant, and low-cost educational robots, our framework succeeds at estimating their 3D configuration from vision alone, discovers their kinematic structure without expert intervention, and executes desired motion trajectories with high precision using a single RGB camera.
For the first time, our system enables modeling and control of 3D-printed, compliant systems without any human modeling and under significant changes in their dynamics, replacing a month-long expert modeling process that nevertheless cannot account for changes in material, dynamics, or manufacturing tolerances.

Our framework allows us to control a wide range of robots from vision alone. For this to be feasible, it is critical that the differential kinematics of the robot can be inferred from vision alone. Some applications of interest may violate this assumption. For instance, when observing mobile legged robots from an external camera, the camera may not observe whether a given leg is touching the ground or not, and thus, cannot determine the motion of the robot as a function of that leg's actuation. Similarly, for dexterous manipulation, sensing contacts with an object is critical. Conditioning the deep-learning based inference method for Neural Jacobian Fields on additional sensors such as tactile ones~\cite{higuera2023neural, zhong2023touching,dou2024tactile} could effectively address this limitation. Though at test time, only a single camera is necessary to control a robot, Neural Jacobian Fields currently requires multi-view video at training time. 

Our method dramatically broadens the design space of robots by decoupling their hardware from their modeling and control. 
We anticipate that our method will enable the deployment of bio-inspired, hybrid soft-rigid robots that were previously practically impossible to model and control. 
Our method further has the potential to lowering the barrier-to-entry to robotic automation by enabling the control of mass-producible, low-cost robots that lack the precision and sensing capabilities to be controlled with conventional methods.
%
%
\section{Methods}
\label{app:methods}

First, we will give a general overview of the data our system ingests.
Then, we will provide details on neural 3D reconstruction and scene representation, as well as their training details via differentiable rendering. 
Next, we will describe mathematical insights from our Jacobian Field parameterization.
Lastly, we will describe the robotic systems used in our paper, including details on their modeling, control, sensing, actuation, morphology, material, fabrication, and costs.

\subsection{Dataset Collection}
Our method is fully self-supervised and does not require any manual data annotation.
We illustrate the data collection process in Extended Data Figure~\ref{fig:fig5_method_overview}a. 
We capture multiple video streams of the robot executing random actions. 
Specifically, we set up 12 consumer-grade cameras that observe the robot from 12 different perspectives.

We obtain intrinsics directly from the cameras. We calibrate camera poses using \SI{3}{\centi\meter} April tags~\cite{olson2011apriltag}.
%
We denote the vector of motor set points as \joints. We first manually select a safe range for each of the command lines.
To create a single data sample, we randomly select from a uniform distribution $u$, execute the command, and wait for it to settle to a steady state. 
We then capture images with all 12 cameras, and denote the timestep as $t$.

We then uniformly sample a change in the motor commands $\delta\joints_{t}$. The next step command %
$\joints_{t+1} = \joints_{t}+\delta\joints_{t}$ 
is then executed on the robot. We again capture images with all 12 cameras, and denote this as timestep $t+1$. 
This leads to a multiview image dataset, $\{(\img_t^0, ..., \img^{11}_t)\}_{t=0}^T$, where the superscript denotes the camera index and $t$ denotes the time step.
While our method does not strictly depend on it, leveraging RGB-D cameras that capture depth in addition to color accelerates training due to additional geometry supervision. Hence, we use Intel RealSense D415 RGB-D cameras for all of the experiments in this paper.
Finally, we extract 2D motion information from this dataset via an off-the-shelf optical flow method, RAFT~\cite{raft}, which takes as input two consecutive video frames of one of the cameras and computes optical flow $\oflow^i_t$ between an image captured at time $t$ and $t+1$ by camera $i$, caused by the motor command $\delta \joints_t$.
%
For each timestep $t$, our training dataset is thus a tuple of the following form:
\begin{equation}
    (\{(\img^0_t, \depth^0_t, \oflow^0_{t}, \pose^0, \intrinsics^0), ..., (\img^{11}_t, \depth^{11}_t, \oflow^{11}_{t }, \pose^{11}, \intrinsics^{11})\}, \delta \joints_t),
    \label{eq:train_tuple}
\end{equation}
i.e., RGB images $\img^i_t$, depth $\depth^i_t$, optical flow $\oflow^i_t$, pose $\pose^i$, and intrinsics $\intrinsics^i$ for the $i$-th out of 12 cameras, as well as the change in robot command $\delta\joints_t$.

\subsection{Neural 3D Reconstruction and Neural Scene Representation}
Given a single image, we leverage deep learning to reconstruct both the proposed Jacobian Field as well as a Neural Radiance Field.
Both the Jacobian Field and the Neural Radiance Field are functions that map a 3D coordinate \coord{} to either the system Jacobian or the radiance and occupancy.

We follow pixelNeRF~\cite{pixelnerf} to reconstruct both these representations.
Given an image $\img \in \mathbb{R}^{\h \times \w \times 3}$ with height $\h$ and width $\w$, we first extract a 2D feature volume $\featvol \in \mathbb{R}^{\h /p \times \w /p \times n}$ where $p$ indicates downsampling resulting from convolutions with stride larger than 1. 
Suppose we want to predict the Jacobian \jacobian, radiance \radiance, and density \density{} at a 3D coordinate \coord.
We first project that 3D coordinate onto the image plane using the known camera calibration as $\pi(\coord)$.
We then sample the feature volume at the resulting pixel coordinate using bilinear interpolation $\featvol(\pi(\coord))$.
We finally predict the Jacobian \jacobian, radiance \radiance, and density \density using a fully connected neural network $\text{FC}$:
\begin{equation}
    (\jacobian, \radiance, \density) = \text{FC}(\featvol(\pi(\coord)), \gamma(\coord)),
 \label{eq:pixelnerf}
\end{equation}
where $\gamma(\coord)$ denotes sine-cosine positional encoding of \coord{} with six exponentially increasing frequencies~\cite{mildenhall2020nerf}.

\subsection{Training via Differentiable Rendering}
We illustrate the training loop of our system in Extended Data Figure~\ref{fig:fig5_method_overview}b.
In each forward pass, we sample a random timestep $t$ and its corresponding training tuple as described in Equation~\ref{eq:train_tuple}.
We then randomly pick two of the twelve cameras and designate one as the source camera and one as the target camera.
The key idea of our training loop is to predict both the image as well as the optical flow observed by the target camera given the input view $\img_\text{input}$ and the robot action $\delta \joints_t$.
Both image and optical flow of the target view are generated from the radiance and Jacobian fields via volume rendering~\cite{mildenhall2020nerf}. Our following discussion closely follows that of pixelNeRF~\cite{pixelnerf}.

We first parameterize the rays that go through each pixel center as $\ray(\volrendidx) = \textbf{o} + \volrendidx \textbf{e}$, with the camera origin $\textbf{o} \in \R^3$ and the ray unit direction vector $\textbf{e} \in \R^3$.
We then use volume rendering to predict RGB $\hat{\mathbf{I}}$ and depth $\hat{\mathbf{d}}$ images:
\begin{gather}
     \hat{\mathbf{I}}(\ray) = \int_{t_n}^{t_f} T(t) \density(t) \radiance(t) \, dt \\
     \hat{\mathbf{d}}(\ray) = \int_{t_n}^{t_f} T(t) \density(t) t \, dt 
 \label{eq:rendering}
\end{gather}
where~$T(t) = \exp(- \int_{t_n}^{t} \density(s) \,ds)$ accounts for occlusion via alpha-compositing, that is, points closer to the camera with a nonzero density $\density$ will occlude those points behind them.
For each ray $\ray$ of the target camera, we then densely sample 3D points between near $t_n$ and far $t_f$ depth bounds. 
For each 3D point $r(t) \in \mathbb{R}^{3}$, we obtain its density $\density$ and color $\radiance$, and Jacobian $\jacobian$ from Equation~\ref{eq:pixelnerf}. 
The notation $\img(\ray)$ selects the pixel in the image $\mathbf{I}$ that corresponds to the ray $\ray$.

Predicted optical flow $\hat{\oflow}(\ray)$ is also computed via volume rendering. For every 3D point along a ray, we use the Jacobian quantity to advect the 3D ray sample via $\ray(t) + \mathbf{J}(t) \delta \mathbf{u}$.
Then, we apply alpha compositing to both original 3D ray samples and their advected counterparts to obtain $\hat{\mathbf{x}}(\ray)$, $\hat{\mathbf{x}}^{+}(\ray) \in \mathbb{R}^{3}$
\begin{gather}
     \hat{\mathbf{x}}(\ray) = \int_{t_n}^{t_f} T(t) \density(t) \ray(t) \, dt \\
     \hat{\mathbf{x}}^{+}(\ray) = \int_{t_n}^{t_f} T(t) \density(t) (\ray(t) + \mathbf{J}(t) \delta \mathbf{u}) \, dt
\end{gather}
Finally, to obtain $\hat{\oflow}(\ray)$, we project $\hat{\mathbf{x}}(\ray)$, $\hat{\mathbf{x}}^{+}(\ray)$ to the 2D image coordinate using camera intrinsic and extrinsic parameters and compute the positional difference
\begin{equation}
    \hat{\oflow}(\ray) = \hat{\mathbf{x}}^{+}(\ray)_{\text{image}} -  \hat{\mathbf{x}}(\ray)_{\text{image}}
    \label{eq:optical_flow_prediction}
\end{equation}

\subsubsection{Supervising the Robot Geometry via RGB-D Renderings.}
To predict the RGB and depth images captured by the target camera, we rely on the radiance field components, color  field \radiance{} and density field \density{}, in Equation~\ref{eq:pixelnerf}.  
The predictions for the RGB image and depth image observed by the target camera are obtained by alpha-compositing the RGB colors and sample depths for each pixel according to Equation~\ref{eq:rendering}.
For each target image with its corresponding pose $\pose$, we compute losses
\begin{gather}
    \mathcal{L}_{\text{RGB}} = 
    \sum_{\ray \in \mathcal{R}}
        \left\lVert
   \hat{\textbf{I}}(\ray)
   -
   \textbf{I}(\ray)
    \right\rVert_2^2, \\
    \mathcal{L}_{\text{depth}} = 
    \sum_{\ray \in \mathcal{R}}
        \left\lVert
   \hat{\textbf{d}}(\ray)
   -
   \textbf{d}(\ray)
    \right\rVert_2^2,
    \label{eq:rendering_loss}
\end{gather}
where $\mathcal{R}$ is the set of all rays in the batch.
Minimizing these losses trains our model to recover the correct density values and thus, the robot geometry. Note that the depth loss is optional and neural radiance fields are generally trained without it~\cite{mildenhall2020nerf,pixelnerf,tewari2023diffusion}, but as consumer-grade RGB-D cameras are readily available, we rely on this additional signal.

\subsubsection{Supervising the Jacobian Field by Predicting 2D Motion.}
%
We compute a 2D \emph{motion loss} using ground truth motion tracks to supervise the Jacobian Field.
\begin{equation}
    \mathcal{L}_{\text{motion}} = 
    \sum_{\ray \in \mathcal{R}}
        \left\lVert
   \hat{\oflow}(\ray)
   -
   \oflow(\ray)
    \right\rVert_2^2.
    \label{eq:rendering_loss}
\end{equation}
Minimizing this loss trains our model to predict the correct system Jacobian at each 3D point.

\subsection{Jacobian Field Details}
\label{sec:jacobian_field_details}

Our Jacobian field is a dense, spatial 3D generalization of the conventional system Jacobian in the context of dynamical systems. In this section, we mathematically describe the motivations and insights of our parameterization.
We first derive the conventional system Jacobian. Consider a dynamical system with state $\mathbf{q} \in \mathbb{R}^{m}$, input command $\mathbf{u} \in \mathbb{R}^{n}$, and dynamics $\mathbf{f}: \mathbb{R}^{m} \times \mathbb{R}^{n} \mapsto \mathbb{R}^{m}$. Upon reaching a steady state, the state of the next timestep $\mathbf{q}^{+}$, is given by
\begin{equation}
    \mathbf{q}^{+} = \mathbf{f}(\mathbf{q}, \mathbf{u}).
    \label{eqn:full_dynamic}
\end{equation}
Local linearization of $\mathbf{f}$ around the nominal point $(\bar{\mathbf{q}}, \bar{\mathbf{u}})$ yields 
\begin{equation}
    \mathbf{q}^{+} = \mathbf{f}(\bar{\mathbf{q}}, \bar{\mathbf{u}}) + \frac{\partial \mathbf{f}(\mathbf{q}, \mathbf{u})}{\partial \mathbf{u}} \bigg\rvert_{\bar{\mathbf{q}}, \bar{\mathbf{u}}} \delta \mathbf{u}.
    \label{eqn:system-linearization}
\end{equation}
Here, $\mathbf{J}(\mathbf{q}, \mathbf{u}) = \frac{\partial \mathbf{f}(\mathbf{q}, \mathbf{u})}{\partial \mathbf{u}}$ is known as the system Jacobian, the matrix that relates a change of command $\mathbf{u}$ to the change of state~$\mathbf{q}$.

Conventionally, modeling a robotic system involves experts designing a state vector $\mathbf{q}$ that completely defines the robot state, and then embedding sensors to measure each of these state variables. For example, the piece-wise-rigid morphology of conventional robotic systems means that the set of all joint angles is a full state description, and these are conventionally measured by an angular sensor in each joint. 
%
However, these design decisions are challenging for soft and hybrid soft-rigid systems. First, instead of discrete joints, large parts of the robot might deform. Embedding sensors to measure the continuous state of a deformable system is difficult, both because there is no canonical choice for sensors universally compatible with different robots and because their placement and installation are challenging. Next, designing the state itself is challenging - in contrast to a piece-wise rigid robot, where the state vector can be a finite-dimensional concatenation of joint angles, the state of a deformable robot is infinite-dimensional due to continuous deformations.  


Our Jacobian Field solves these challenges. First, the combination of Jacobian and Neural Radiance Fields is a complete representation of the robot state - it encodes the position of every 3D point of the robot, as well as its kinematics, i.e., how that 3D point would move under any possible action. This absolves us from the need to manually model a robot state $\mathbf{q}$. 
Second, we note that for many robotic systems, it is possible to infer their 3D configuration from vision alone. Even if parts of the robot are occluded, we are often still able to infer their 3D position from the visible parts of the robot - just like observing the back of a human arm allows us to infer what the occluded side will look like. In this work, we infer the state completely from a single camera, but it is straightforward to add additional cameras to achieve better coverage of the robot~\cite{pixelnerf}.

We now derive the connection of the Jacobian Field and the per-camera 2D optical flow we use for its supervision.
Rearranging Equation~\ref{eqn:system-linearization} yields 
\begin{equation}
    \mathbf{q}^{+} - \mathbf{f}(\bar{\mathbf{q}}, \bar{\mathbf{u}}) = \frac{\partial \mathbf{f}(\mathbf{q}, \mathbf{u})}{\partial \mathbf{u}} \bigg\rvert_{\bar{\mathbf{q}}, \bar{\mathbf{u}}} \delta \mathbf{u}.
\end{equation}
In practice, our nominal point represents a steady state, as one can wait for the robot command to settle. Then, $\mathbf{f}(\bar{\mathbf{q}}, \bar{\mathbf{u}})$ is approximately $\bar{\mathbf{q}}$. We consolidate $\delta \mathbf{q} = \mathbf{q}^{+} - \bar{\mathbf{q}}$ to express the \emph{change in robot state} $\delta \mathbf{q}$ as a function of the System Jacobian
\begin{equation}
    \delta \mathbf{q} = \frac{\partial \mathbf{f}(\mathbf{q}, \mathbf{u})}{\partial \mathbf{u}} \bigg\rvert_{\bar{\mathbf{q}}, \bar{\mathbf{u}}} \delta \mathbf{u}.
\end{equation}
We define the dense 3D position of every robot point as the state of the robot. Consequently, the change in robot state $\delta \mathbf{q}$ can be understood as the 3D velocity field that moves these points according to the action $\delta \mathbf{u}$. 
The 3D velocity field $\delta \mathbf{q}$ can be measured as 2D pixel motions $\mathbf{v}^i$ across all camera views using off-the-shelf optical flow and point-tracking methods. 
Given a training data sample $(
\oflow^i, \delta \mathbf{u}, \intrinsics^{i}, \pose^{i})$, our neural Jacobian field $\mathbf{J}(\mathbf{x}, I)$ associates the two signals ($\delta \mathbf{q}$, $\delta \mathbf{u}$) via $\widehat{\oflow}^i = \text{render}(\mathbf{J}(\mathbf{X}, \img), \delta \mathbf{u}, \intrinsics^{i}, \pose^{i})$ using Equation~\ref{eq:optical_flow_prediction}, where $\mathbf{X}$ represents samples of 3D coordinates from the neural field.
%



To sum up, our Jacobian Field leverages visual motion measurements as a learning signal and can be trained self-supervised purely by observing robot motion under random actions with multi-view cameras.
It directly relates the change in robot state $\delta \mathbf{q}$, defined as the 3D motion field that advects every 3D point of the robot according to the action $\delta \mathbf{u}$, to the motion in 2D-pixel space observed by multiple cameras. This provides a signal for learning which part of 3D space is sensitive to a particular command of the robotic system, and enables control by specifying the desired motion of any robot point in 2D or 3D. 
%
%
\subsection{Domain Randomization Details}
We apply domain randomization techniques~\cite{tobin2017domain, peng2018sim, florence2018dense} to the input images during training.
This process trains our framework to be robust against occlusion, background changes, and other visual distribution shifts. 
For every input frame collected in our video dataset, we apply a motion threshold to the point tracks and use the Segment Anything framework~\cite{kirillov2023segment} to obtain a binary mask corresponding to the robot.
We apply background domain randomization as described and implemented in~\cite{florence2018dense}. 
We apply foreground domain randomization by overlaying natural images randomly sampled from the coco dataset~\cite{lin2014microsoft} on our training video frames.
Before overlaying, we randomly crop the coco images and resize them to be smaller than the dataset video frame size. 
We highlight that the target images used to create supervision are not augmented, and are the undisturbed original observations $\img, \depth, \oflow$. 
This trains the neural single-image-to-3D module to be robust against background changes and partial occlusion. As empirically observed in Extended Data Figure~\ref{fig:fig8_robustness_against_occlusion} and quantitatively tested in Extended Data Figure~\ref{fig:fig6_sensitivity_analysis}, our framework's depth and Jacobian predictions are robust against visual perturbations.


\subsection{Robot Systems}

\subsubsection{Pneumatic Hand}
Based on a design first introduced in~\cite{matusik2023directly}, the pneumatically actuated soft robot hand is 3D printed in one piece using vision-controlled jetting~\cite{buchner2023vision}. This printing technique is based on inkjet deposition and enables the combination of soft and rigid materials. The fingers are based on PneuNet actuators~\cite{ilievski2011soft}, and the palm features a rigid core surrounded by soft elastomer skin. The hand does not require manual assembly after printing, and the total fabrication cost is around \$300. The hand is driven by a 15-channel proportional valve terminal (MPA-FB-VI, Festo). Each channel can be individually controlled to adjust the pressure in each of the 15 degrees of freedom (DoF) of the hand. 
For our experiments, we employ two versions of this hand. One is marked by a pen with blue crosses and operated standing upright in the workspace. The other is unmarked and mounted to the tool flange of an industrial robot arm (UR5, Universal Robots A/S). We use the two shoulder joints of the arm to move the hand horizontally and vertically in the workspace.

\subsubsection{Allegro Hand}
The Allegro hand (Wonik Robotics Co. Ltd.) is a commercially available, 16-DoF anthropomorphic robot hand with four fingers~\cite{allegrohand}. Each of the fingers is actuated by four servos and has soft silicone padding at the fingertips. The servos provide joint position feedback and can be torque-controlled via current control. The Allegro hand is popular among researchers across academia and industry, but also expensive, with a price tag upwards of \$15,000~\cite{lambeta2020digit, shaw2023leap, funabashi2020stable, zhu2019dexterous}.

\subsubsection{HSA Platform}
The HSA Platform is a servo-driven, 4-DoF soft robotic platform based on compliant actuators made out of handed shearing auxetics (HSAs). Introduced in~\cite{lipton2018handedness}, HSAs are metamaterials with patterns that couple extension with shearing. Tiled on the surface of a cylinder, HSAs enable linear actuators that extend when twisted. These actuators are 3D printed from polyurethane resins (FPU 50, Carbon Inc.) via digital light projection, according to~\cite{truby2021recipe}. Four actuators are arranged in a 2\texttimes2 configuration, joined rigidly together at the top, and each driven by a servomotor at the bottom. The resulting robotic platform has three rotational DoFs that mimic the movements of a human wrist and one translational DoF that allows it to extend and contract. The HSA platform costs about \$1,200 to make, with the four servos (Dynamixel MX-28, ROBOTIS) at \$260 each dominating the cost. Even though a model that relates servo position feedback to platform pose can be learned, the relationship breaks down when the compliant HSA platform deforms in response to external forces~\cite{zhang2023machine}. HSA-based robots can be sensorized via fluidic innervation~\cite{truby2022fluidic} and embedding internal cameras~\cite{zhang2022vision}. However, each of these methods comes at the price of significantly increased fabrication complexity.

\subsubsection{Poppy Ergo Jr}
The Ergo Jr is an ultra-low-cost 6-DoF robot arm for educational purposes and part of the DIY open-source robotics platform called the \textit{Poppy Project}~\cite{poppyergojr, lapeyre2015poppy}. The Ergo Jr is driven by six low-cost servos (Dynamixel XL-320, ROBOTIS) and can be assembled in 30-45 minutes~\cite{poppydoc}. Apart from the servos and the electronics, its parts are 3D printed on a consumer-grade desktop FDM printer (P1S, Bambu Lab). The Ergo Jr cost us \$270 in total (Extended Data Table~\ref{tab:table4_poppy_cost}), which can be reduced further if the servo control board was custom-made instead of bought off-the-shelf. The Ergo Jr is not equipped with any sensors other than the servo encoder. Past research~\cite{salvato2021characterization, chevalier2019robo, golemo2018sim} describes this robot as difficult to model due to its low manufacturing quality and significant backlash in its kinematic chain.

\section{Data Availability}
All data needed to evaluate the conclusions in the paper are present in the paper or the extended data. The training data will be publicly released.

\section{Code Availability}
The full source code for training the Neural Jacobian Fields, deploying them on the robots, and reproducing the results will be made available on GitHub at \url{https://github.com/sizhe-li/neural-jacobian-field}

\section{Supplementary Information}
\textbf{Supplementary Video 1.} Summary of the results. 
3D reconstruction of Jacobian Fields for all robots, 3D motion predictions for all robots, visuomotor control of all robots using their Jacobian Fields.

\section{Acknowledgements}
The authors thank Hyung Ju Terry Suh for his writing suggestions (system dynamics) and Tao Chen and Pulkit Agrawal for their hardware support on the Allegro hand. 
V.S. acknowledges support from the Solomon Buchsbaum Research Fund through MIT's Research Suppport Committee. S.L.L. was supported through an MIT Presidential Fellowship.
A.Z., H.M., C.L., and D.R. acknowledge support from the National Science Foundation EFRI grant 1830901 and the Gwangju Institute of Science and Technology.

\section{Author Contributions}
A contribution to the main ideas and methods of the work was made by S.L.L. (neural jacobian field, applications to soft robots) and V.S. (neural jacobian field, applications to soft robots).
A contribution to the hardware setup was made by A.Z. (HSA platform, Poppy robot arm, camera setup), C.L. (pneumatic hand, FESTO system), B.C. (Poppy robot arm, camera setup), S.L.L. (Allegro Hand, camera setup), H.M. (pneumatic hand).
A contribution to the formalization of modeling and control challenges in soft robots was made by A.Z., C.L., D.R.
A contribution to experiment design was made by A.Z. (robustness studies, pneumatic hand), C.L. (pneumatic hand), S.L.L. (all experiments), V.S. (all experiments).
A contribution to the implementation and experimentation of the Neural Jacobian Field was made by S.L.L.
A contribution to the data analysis and interpretation was made by S.L.L. (analysis, interpretation), A.Z. (interpretation) and V.S (interpretation)
A contribution to compute resources, funding, hardware, and facilities was made by V.S. (compute hardware, funding, facilities) and D.R. (robot hardware, funding, facilities).
S.L.L., A.Z., B.C., C.L., V.S. drafted the work. S.L.L., A.Z., V.S. substantively revised the work. All authors approved the final draft of the manuscript.

\section{Competing Interests}
The authors declare no competing interests.


\printbibliography

\appendix
\newpage
\section{Extended Data}\label{app:extended}
\begin{figure}[!h]
\centering
\includegraphics[width=\textwidth]{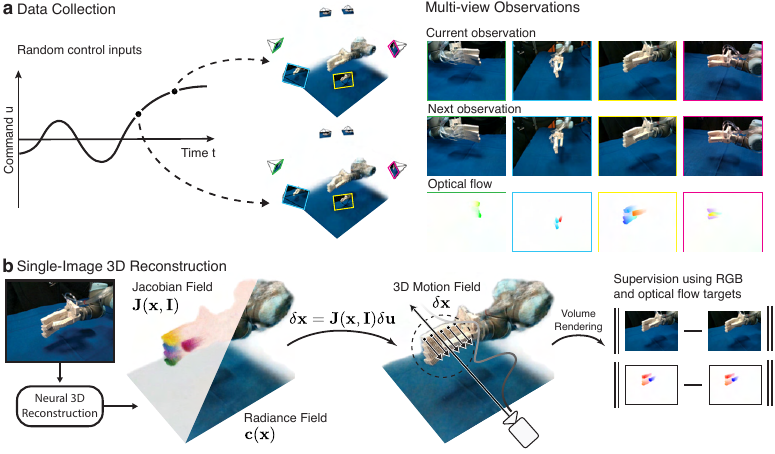}
\caption{
\textbf{Overview of dataset collection, training, and inference processes.} 
\textbf{a}, 
Our data collection process samples random control commands to be executed on the robot. Using a setup of 12 RGB-D cameras, we record multi-view captures before each command is executed, and after each command has settled to the steady state.
%
\textbf{b}, our method first conducts neural 3D reconstruction that takes a single RGB image observation as input and outputs the Jacobian field and Radiance field. 
Given a robot command, we compute the 3D motion field using the Jacobian field. 
Our framework can be trained with full self-supervision by rendering the motion field into optical flow images and the radiance field into RGB-D images.
}
\label{fig:fig5_method_overview}
\end{figure}
\begin{figure}[!h]
\centering
\includegraphics[width=\textwidth]{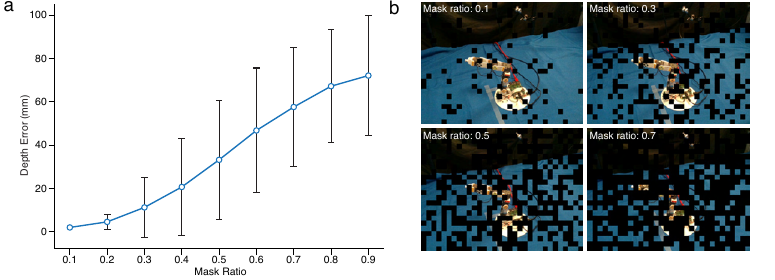}
\caption{
\textbf{Sensitivity analysis of depth prediction.} \textbf{a,} We evaluate the depth reconstruction quality under out-of-distribution scenarios. Given a collection of testing images, we perturb these images with black patches sampled at random locations that achieve increasingly larger mask ratios. We plot the mean depth error over the mask ratio and visualize the standard deviation as the error bar. \textbf{b,} Overlaying black patches on an example input image with increasingly larger mask ratios.  
}
\label{fig:fig6_sensitivity_analysis}
\end{figure}
\begin{figure}[!h]
\centering
\includegraphics[width=\textwidth]{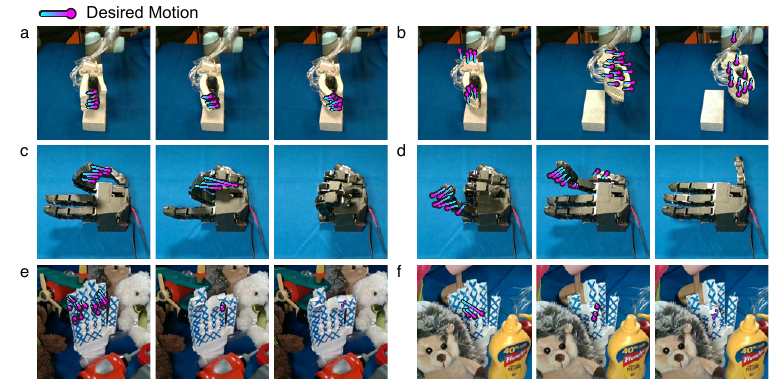}
\caption{
\textbf{Additional evaluation on visuomotor control.} 
\textbf{(a, b)}, Our framework controls a 3D-printed pneumatic hand to grasp on an object and complete a pick-and-place task.
\textbf{(c, d)}, Our approach controls the Allegro hand to close and open each finger fully.
\textbf{(e, f)}, We intentionally perturb the scenes to create occlusion and background changes. We find that Neural Jacobian Field is robust against out-of-distribution scenarios and successfully controls the hand to achieve detailed motions.
}
\label{fig:fig7_more_visual_result}
\end{figure}
\begin{figure}[!h]
\centering
\includegraphics[width=\textwidth]{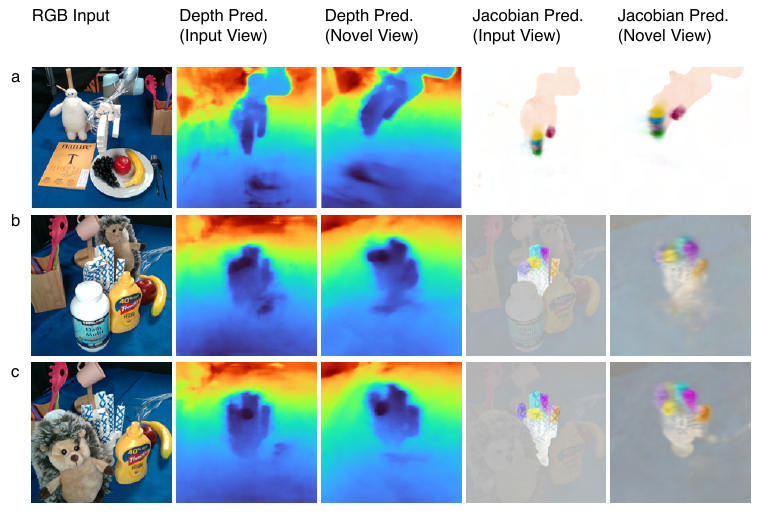}
\caption{
\textbf{Qualitative results on robustness against out-of-distribution scenarios.}
\textbf{a}, We perturb the visual scenes by placing a dozen objects around the pneumatic hand mounted on the robot arm. Consequently, the input observation is highly out-of-distribution from the training data.
We visualize the predictions of depth and Jacobian at both the input and novel viewpoints. We find that our method retains high-quality predictions.
\textbf{(b, c)}, We placed a dozen objects around the pneumatic hand to create occlusion. For presentation clarity, we overlay the Jacobian prediction on RGB images to highlight the shapes of the hand with masking.
}
\label{fig:fig8_robustness_against_occlusion}
\end{figure}

\begin{table*}[!h]
\centering

\small

\resizebox{\textwidth}{!}{
\begin{tabular}{@{}c|cc|cccc|c|c@{}}
\toprule
 & \multicolumn{2}{c|}{\textbf{a. Allegro Hand}} & \multicolumn{4}{c|}{\textbf{b. HSA Platform}} & \textbf{c. Poppy Arm} & \textbf{d. Pneumatic Hand} \\ \midrule
     & Angle error & Location error & \multicolumn{4}{c|}{Marker error in millimeter}     & Marker error  & Pressure error \\
     & in degree   & in millimeter  & w/o weight & w/ weight & rotational & translational & in millimeter & in millibar    \\ \midrule
Mean & 2.568       & 3.667          & 5.354      & 7.784     & 7.911      & 3.372         & 5.339         & 5.025          \\
Std  & 1.595       & 3.297          & 2.343      & 0.481     & 0.398      & 0.616         & 1.735         & 2.161          \\ \bottomrule
\end{tabular}
}
\caption{
\textbf{Quantitative evaluation on visuomotor control.} 
\textbf{a-d}, We measure the distance between the final state achieved by our method and the ground truth desired state.
\textbf{a}, Joint errors in degree angle measurement and finger-tip errors in positional measurement. 
\textbf{b}, Motion capture marker location error between the final state achieved by our method and the ground truth recorded in the reference trajectory.
\textbf{c}, Similar to b., we report errors computed based on the locations of the motion capture markers.
\textbf{d}, Pressure errors measured in millibar from the 15-channel proportional valve terminal.
}
\label{tab:table1_control_result}
\end{table*}

\begin{table*}[!h]
\centering
\large
\resizebox{\textwidth}{!}{
\begin{tabular}{@{}c|cc|cc|cc|cc@{}}
\toprule
 &
  \multicolumn{2}{c|}{\textbf{Allegro Hand}} &
  \multicolumn{2}{c|}{\textbf{Handed Shearing Auxetics}} &
  \multicolumn{2}{c|}{\textbf{Poppy Toy Robot Arm}} &
  \multicolumn{2}{c}{\textbf{Pneumatic Hand}} \\ \cmidrule(l){2-9} 
     & Depth error (mm) & Flow error (pix.) & Depth error (mm) & Flow Error (pix.) & Depth error (mm) & Flow error (pix.) & Depth error (mm) & Flow error (pix.) \\ \midrule
Mean & 5.033            & 1.305      & 1.109            & 3.597      & 1.206            & 6.797      & 6.519            & 1.150      \\
Std  & 3.080            & 1.504      & 0.576            & 2.779     & 0.619            & 7.287      & 2.028            & 1.520      \\ \bottomrule
\end{tabular}
}

\caption{
\textbf{Quantitative evaluation on 3D reconstruction.} Using testing images from all camera viewpoints, we measure depth prediction errors in millimeters and optical flow prediction errors on a $640\times480$ image grid.
The depth errors are computed as the L2 distance between the ground truth depth value measured by the Intel Realsense D415 RGB-D cameras and the predicted depth value averaged over all pixel locations in the image.
The optical flow errors are computed as the L2 distance between the ground truth measured by the point tracker~\cite{doersch2023tapir} and the prediction averaged over all pixel locations in the image.
}
\label{tab:table2_perception_result}
\end{table*}

\begin{table}[!h]
\begin{tabular}{lcrr}
\hline
\multicolumn{1}{c}{\textbf{Part}}                         & \textbf{Quantity} & \multicolumn{1}{c}{\textbf{Unit Cost}} & \multicolumn{1}{c}{\textbf{Total}} \\ \hline
ROBOTIS Dyanmixel XL-320 servos                           & 6                 & \$           26.90                     & \$        161.40                   \\
Pixl Poppy Ergo Jr servo control board                    & 1                 & \$           32.47                     & \$           32.47                 \\
ROBOTIS RS-10 rivets                                      & 1                 & \$              6.60                   & \$              6.60               \\
ROBOTIS rivet tool                                        & 1                 & \$              1.10                   & \$              1.10               \\
Raspberi Pi 3 Model B+ board                              & 1                 & \$           48.99                     & \$           48.99                 \\
Gigastone 8GB microSD card                                & 1                 & \$              3.30                   & \$              3.30               \\
PwrON 7.5V 2A AC DC power supply                          & 1                 & \$              9.99                   & \$              9.99               \\
Vabogu-CAT8 ethernet cable                                & 1                 & \$              3.99                   & \$              3.99               \\
Bambu Lab PLA filament for 3D printed parts     & 108 g                 & \$              20 per kg                   & \$              2.16               \\
Screws, nuts, standoffs (estimate)                                         &                   &                                        & \$              2.00               \\ \hline
\textbf{Grand Total}                                      &                   &                                        & \textbf{\$        272.00}         
\end{tabular}
\caption{
\textbf{Bill of materials for Poppy Ergo Jr arm.}
}
\label{tab:table4_poppy_cost}
\end{table}











\end{document}